\documentclass[authorversion, format=sigconf]{acmart}

\usepackage[utf8]{inputenc}
\usepackage{eqnarray}
\usepackage{algorithm}
\usepackage[capitalise,noabbrev]{cleveref}

\usepackage{enumitem}
\usepackage[noend]{algpseudocode}
\usepackage{amsmath}
\usepackage{xspace}
\usepackage{mathtools}
\usepackage{appendix}
\usepackage{microtype}

\usepackage{xcolor}
\usepackage{booktabs} % For formal tables
\usepackage{makecell}

\newcommand{\momex}{\textsc{\mome -x}\xspace}
\newcommand{\mome}{\textsc{mome}\xspace}
\newcommand{\pgx}{\textsc{\mome -pgx}\xspace}
\newcommand{\mesum}{\textsc{me-sum}\xspace}
\newcommand{\meenergy}{\textsc{me-stability}\xspace}
\newcommand{\memagnetism}{\textsc{me-magnetism}\xspace}

\newcommand{\mapelites}{\textsc{map-elites}\xspace}

\newcommand{\moqd}{\textsc{moqd}\xspace}

\newcommand{\qd}{\textsc{qd}\xspace}

\newcommand{\ga}{\textsc{ga}\xspace}

\newcommand{\csp}{\textsc{CSP}\xspace}

\newcommand{\replications}{\textsc{15}\xspace}

\newcommand{\qdscore}{\textsc{qd-score}\xspace}
\newcommand{\energyqdscore}{\textsc{energy-qd-score}\xspace}
\newcommand{\magmomqdscore}{\textsc{magnetism-qd-score}\xspace}
\newcommand{\moqdscore}{\textsc{moqd-score}\xspace}

\newcommand{\globalhypscore}{\textsc{global-hypervolume}\xspace}
\newcommand{\coverage}{\textsc{coverage}\xspace}

\newcommand{\code}[0]{\url{https://github.com/adaptive-intelligent-robotics/MOQD-CSP}}

\copyrightyear{2024}
\acmYear{2024}
\setcopyright{rightsretained}
\acmConference[GECCO '24]{Genetic and Evolutionary Computation Conference}{July 14--18, 2024}{Melbourne, VIC, Australia}
\acmBooktitle{Genetic and Evolutionary Computation Conference (GECCO '24), July 14--18, 2024, Melbourne, VIC, Australia}
\acmDOI{10.1145/3638529.3654048}
\acmISBN{979-8-4007-0494-9/24/07}

\begin{document}

\title{Multi-Objective Quality-Diversity for Crystal Structure Prediction}

\renewcommand{\shorttitle}{Multi-Objective Quality-Diversity for Crystal Structure Prediction}

\author{Hannah Janmohamed}
\affiliation{%
  \institution{Imperial College London, InstaDeep}
  \city{London}
  \country{UK}}
\email{hnj21@imperial.ac.uk}

\author{Marta Wolinska}
\affiliation{%
 \institution{Imperial College London}
  \city{London}
  \country{UK}}
% \email{}

\author{Shikha Surana}
\affiliation{%
 \institution{InstaDeep}
  \city{London}
  \country{UK}}
% \email{s.surana@instadeep.com}

\author{Thomas Pierrot}
\affiliation{%
 \institution{InstaDeep}
  \city{Boston}
  \country{USA}}
% \email{t.pierrot@instadeep.com}

\author{Aron Walsh}
\affiliation{%
 \institution{Imperial College London}
  \city{London}
  \country{UK}}
% \email{}

\author{Antoine Cully}
\affiliation{%
  \institution{Imperial College London}
  \city{London}
  \country{UK}}
% \email{a.cully@imperial.ac.uk}

\renewcommand{\shortauthors}{Janmohamed, et al.}

\begin{abstract}
%% one or two sentences providing a basic introduction to the field
Crystal structures are indispensable across various domains, from batteries to solar cells, and extensive research has been dedicated to predicting their properties based on their atomic configurations.
%% two to three sentences of more detailed background, for scientists in related discplines
However, prevailing Crystal Structure Prediction methods focus on identifying the most stable solutions that lie at the global minimum of the energy function.
This approach overlooks other potentially interesting materials that lie in neighbouring local minima and have different material properties such as conductivity or resistance to deformation.
% One sentence clearly stating the general problem being addressed by this stufy
By contrast, Quality-Diversity algorithms provide a promising avenue for Crystal Structure Prediction as they aim to find a collection of high-performing solutions that have diverse characteristics.
However, it may also be valuable to optimise for the stability of crystal structures alongside other objectives such as magnetism or thermoelectric efficiency.
Therefore, in this work, we harness the power of Multi-Objective Quality-Diversity algorithms in order to find crystal structures which have diverse features \textit{and} achieve different trade-offs of objectives.
% one sentence summarising the main results "here we show..."
We analyse our approach on 5 crystal systems and demonstrate that it is not only able to re-discover known real-life structures, but also find promising new ones.
% two or three sentences explain what the main result reveals in direct comparison to what was thought to be the case previously or how the main results add to previous knowledge
Moreover, we propose a method for illuminating the objective space to gain an understanding of what trade-offs can be achieved.
%% one or two sentences to put the results in a more general context
\end{abstract}




\maketitle

\section{Introduction}

\begin{figure}
    \centering
    \includegraphics[width=\linewidth]{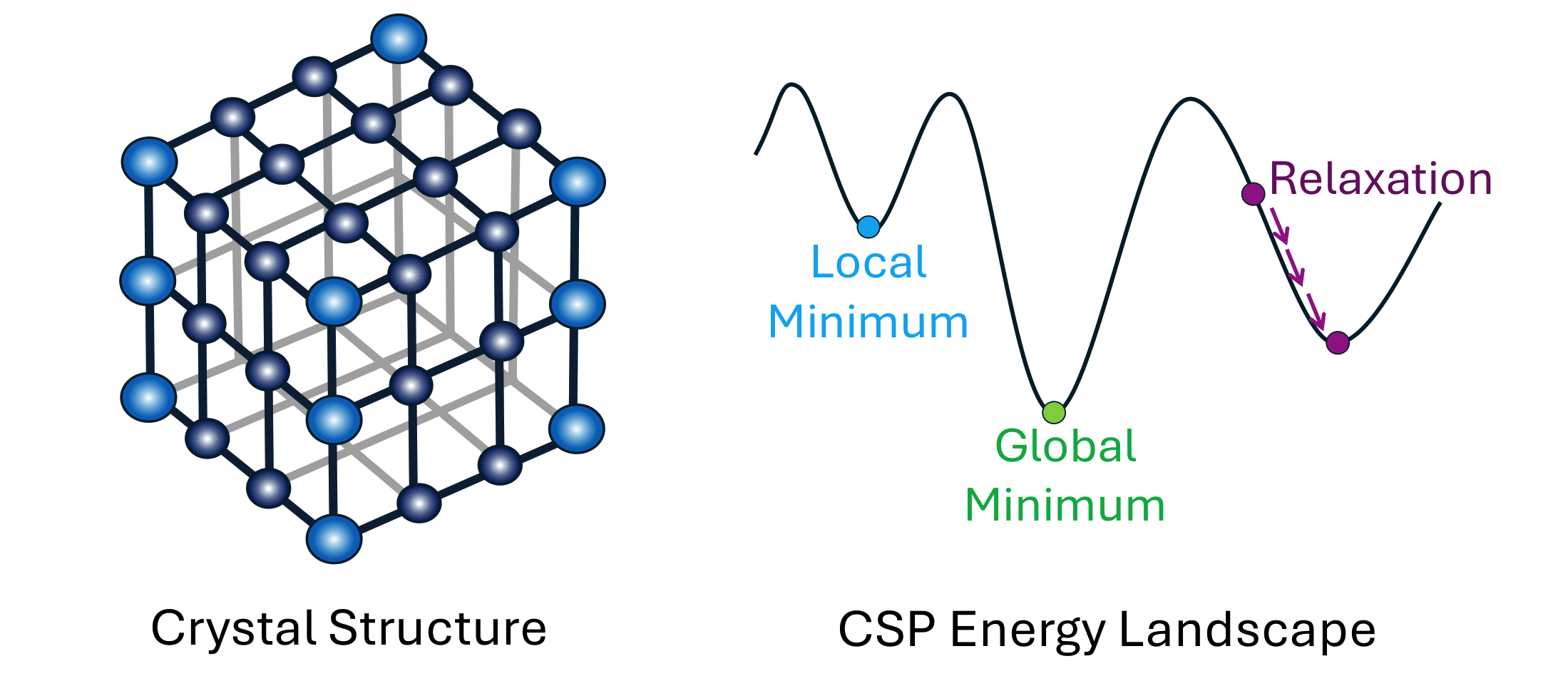}
    \caption{\textit{Left:} Illustration of crystal structure: atoms and molecules are arranged in a repeated, ordered pattern. \textit{Right:} The crystal structure energy landscape is rugged with several local optima. Relaxation is a form of local optimisation that brings solutions toward local optima.}
    \label{fig:csp}
\end{figure}

% "From microchips to batteries and photovoltaics, discovery of inorganic crystals has been bottlenecked by expensive trial-and-error approaches" \cite{gnome}

%% What are crystals, why are they important
Inorganic crystal structures are pervasive in our everyday lives, forming the foundation of materials used in industries as diverse as electronics, pharmaceuticals, and energy storage.
Consequently, advancements in materials science could have revolutionary effects across a wide variety of sectors.
For instance, in the electronics industry, the discovery of novel crystal structures could lead to the development of more efficient semiconductors and superconductors, enabling faster and more energy-efficient electronic devices \cite{gnome}.
% In the pharmaceutical field, understanding the crystal structures of drugs can impact their solubility, bioavailability, and ultimately, their effectiveness in treating various medical conditions.

%% What are existing approaches and why aren't they adequate
    
Crystal Structure Prediction (\csp) is an optimisation problem that aims to predict stable atomic arrangements within crystalline materials.
Traditionally, computational techniques for CSP have focused on finding the global minimum of the energy function, assuming that the crystal structure with the lowest energy is the most stable.
Existing methods, such as genetic algorithms \cite{oganov2011evolutionary}, random search \cite{cheng2022crystal}, and particle swarm optimization \cite{clerc2010particle}, employ various search strategies to locate this global minimum. % Bayesian optimization \cite{pelikan1999boa}
While these methods have made significant contributions to the field, they often overlook potentially valuable regions of the objective space and may miss relevant meta-stable structures in local optima \cite{uspex}.

Consider carbon as an illustrative example.
The most stable carbon crystal is graphite, characterised by its layered hexagonal structure.
However, there are numerous alternative carbon allotropes, such as diamonds, nanotubes, and fullerenes, each possessing distinct properties with immense technological potential.
Traditional \csp methods often fail to explore these alternative structures, as they are solely aimed at finding the lowest energy state.
% \textcolor{red}{Indeed, the \uspex algorithm, which is one of the seminal works in \csp, the authors even note that their $MgSiO_3$ which find perovskite and post-perovskite in a single optimisation run, but the process discarded perovskite because post-perovskite has lower energy \cite{uspex}.}

%% What is QD and why is it better suited, given the problem of previous appraoches
Quality-Diversity (\qd) algorithms \cite{qdunifying}, primarily used in evolutionary robotics and optimization \cite{nature}, have recently emerged as a promising approach for addressing \csp challenges \cite{qd4csp}.
\qd algorithms aim to discover not only the single best solution but instead a large collection of different, yet high-quality, solutions across the search space.
This approach has many advantages in the context of \csp.
Firstly, seeking a collection of solutions could lead to the discovery of multiple local optima, each potentially representing unique and interesting materials.
Discovering a multitude of crystal structures not only enhances our understanding of materials but could also provide a valuable reservoir of options for synthesis. 
In particular, synthesising materials under laboratory conditions can often prove challenging \cite{gnome}, so having alternative candidate solutions offers researchers greater flexibility and choice. 
Finally, since the energy landscape of \csp is typically rugged \cite{uspex}, having a collection of solutions could aid exploration, as diverse solutions can serve as stepping stones to navigate and escape local optima effectively \cite{qdsteppingstones, qdprovably}.

%% Marta paper
Recently, Wolinska et al. \cite{qd4csp} applied the well-established \mapelites \cite{mapelites} algorithm to the \csp problem by using domain-specific mutation operators and graph neural-networks as surrogate models for evaluating potential crystal structures. 
The authors demonstrate that, by using this approach, \qd algorithms are capable of discovering a wide variety of crystal structures \cite{qd4csp}.
Excitingly, their method was not only able to discover structures known by material scientists, but also to  discover some promising new structures.
% Herein lies the motivation for our paper, which introduces a novel approach utilizing multi-objective Quality-Diversity algorithms in the context of CSP
% Our primary contribution is to demonstrate the proof of concept that these algorithms can efficiently uncover a diverse range of stable crystal structures, extending beyond the traditional pursuit of the global energy minimum.
% Through this approach, we not only illuminate the objective space of crystal structures but also provide valuable insights into known stable configurations and the potential discovery of promising, hitherto unexplored solutions.
% In doing so, we offer a unique perspective on crystal structure prediction that addresses the limitations of existing methodologies and paves the way for new frontiers in materials science.

%% What is our method, what is the key contribution and results
While this is an important advancement, using stability as the sole objective does not allow optimisation of any other desired objectives of the materials, such as magnetism or thermoelectric efficiency.
In order to find materials that are useful for downstream applications, it would be extremely valuable to allow for the optimisation of crystal structure stability in tandem with other objectives.
With this goal in mind, in this work, we apply Multi-Objective Quality-Diversity algorithms to the field of \csp.
Multi-objective Quality-Diversity (\moqd) \cite{mome, mome-pgx, c-mome} is a new research effort that aims to find diverse solutions that simultaneously optimise multiple objectives. 
For \csp this allows us to not only find diverse crystalline materials but also explore trade-offs and properties associated with these crystal structures.

%% What is our method
In this paper, we build upon the work of Wolinska et. al \cite{qd4csp} by applying Multi-Objective Quality-Diversity algorithms to Crystal Structure Prediction.
We evaluate our method on 5 different crystal systems and demonstrate that our method is able to find a large collection of crystal structures that have diverse conductivity properties and deformation-resistance, and that achieve varying trade-offs between stability and magnetism.
Moreover, we demonstrate that our approach is not only able to re-discover known structures, but also uncover novel structures that may surpass these known materials.
Finally, we illustrate how our approach can be used to illuminate the search space, highlighting the possible trade-offs and properties that crystal structures can exhibit.
All of our work is fully containerised and is available at: \code.

%% Summary
% In summary, this paper presents a novel application of \moqd algorithms to the \csp problem, offering a promising avenue for materials scientists and researchers to explore and discover stable crystal structures.

% In MendS, found that optimisation over many materials for certain properties returned different structures of the same material that presented different trade-offs of objectives. Therefore, could optimise over one materials and still get different trade-offs\cite{mends}. Moreover, doing so will help to illuminate the objective space w.r.t to material properties which could provide useful insight into the optimisation landscape \cite{mapelites}. 

% Motivation of QD for CSP:
% \begin{itemize}
%     \item "Besides identifying stable phases, this method can thus be used for materials design, both in finding promising structures to synthesize and in giving information on what conditions would be best suited for synthesis" \cite{uspex} (The latter is because USPEX uses constraints which reflect the conditions of the environment in the optimisation process e.g. pressure and temperatures.
% \end{itemize}
%

\section{Background}

\subsection{Crystal Structures}

%% What is a crystal
Crystals are three-dimensional arrangements of atoms or molecules, characterised by a repeating pattern (see \Cref{fig:csp}) \cite{tilley2020crystals}.
The entire crystal structure can be summarised by its \textit{unit cell}, which is the smallest repeating structural component of the crystal \cite{o2020crystal}.
The positions of atoms in the unit cell collectively define the crystal's stability and properties.

% To predict and understand these crystal structures, researchers often rely on computational techniques.
Finding novel crystal structures experimentally in a laboratory setting can be a technically challenging, resource-intensive and time-consuming task.
To overcome these challenges, computational methods have become an indispensable tool for finding new materials.
These techniques leverage the power of computers to explore the vast search space of possible crystal structures, assessing their stability through an \textit{energy function} (see \Cref{fig:csp}).
The energy function calculates the total energy of a given atomic arrangement within a crystal, taking into account the interactions between atoms, including their bond lengths and angles.
In essence, it quantifies the stability of a crystal structure based on the relative positions of its constituent atoms.
The aim of Crystal Structure Prediction (\csp) is to find structures which lie at the global minimum of this energy function and are therefore highly stable.

%% More about the energy function
Computing the energy function of a crystal structure is not always a straightforward task.
In most applications, to accurately predict crystal structures, researchers often employ Density Functional Theory (DFT) calculations \cite{dft}.
DFT calculations estimate the properties of materials based on quantum mechanical principles and, while they are highly accurate, they are also computationally expensive, requiring significant computational resources and time.
To address this, surrogate models \cite{chgnet} have emerged as an alternative approach for approximating the energy function.
These surrogate models aim to reduce the computational burden by providing quick and cost-effective approximations of the energy landscape.
While surrogate models offer efficiency gains, they introduce the challenge of accuracy and reliability, as their predictions may not always align perfectly with the true energy values.

%% Why is CSP hard
The \csp problem is inherently challenging due to the vast high-dimensional space of possible atomic configuration and ruggedness of the energy landscape \cite{uspex}.
Given the complexities of both experimental and computational approaches, crystal structure prediction remains a challenging and critical problem.

\subsection{Quality-Diversity Algorithms}\label{section:qd}

Traditional optimisation methods aim to find a single, high-quality solution, in contrast, the objective of Quality-Diversity (\qd) optimisation algorithms is to find a set of solutions that are both diverse and high-performing. 
In \qd, a solution is characterised by a \textit{fitness} score $f$ and a \textit{feature} vector $d$.
The feature vector characterises some aspect of interest of the solution.
For example, in robotic locomotion the feature vector could reflect the gait of the robot \cite{nature, mome} or in video game level design this could be the number of enemy characters \cite{videogames}.
%This vector points to a specific location within the feature space which represents the domain of features that are relevant or of interest.
The objective of \qd is to find a collection of solutions which have diverse feature vectors and high fitness.

% Additionally, there are two performance criteria that can also be used to evaluate the method: (1) coverage of the feature space and (2) uniformity of the coverage.

In the family of \qd optimisation methods, Multi-dimensional Archive of Phenotypic Elites (\mapelites) is a simple yet powerful algorithm used successfully in several prior works \cite{mapelites, nature, multi-emitter}. 
In \mapelites, the feature space is discretised into a grid composed of cells.
Each cell corresponds to a feature vector and stores the solution characterised by the same feature. 
The algorithm begins with initialising the grid and placing randomly sampled solutions in the cells. 
At each iteration of the \mapelites algorithm, the first step consists of selecting a batch of solutions from the grid. 
Next, copies of these solutions undergo variation to form offspring solutions, which are evaluated to obtain the fitness scores and the features. 
Finally, each new solution is added back to the grid if the cell matching the solution's feature vector is empty or if the existing solution within the cell is outperformed by the new solution.
By repeating this loop for a specified budget, the grid gradually accumulates diverse and high-quality solutions.
The performance of \mapelites algorithms can be assessed via the \qdscore which is the sum of the fitness of the solutions in each cell.

% add to first paragraph, refine examples + add citations 
% QD has been successfully used in robot locomotion, autonomous skill discovery, and procedural content generation.

\subsection{Multi-Objective Optimisation}

\begin{figure}[h!]
    \centering
    \includegraphics[width=0.5\linewidth]{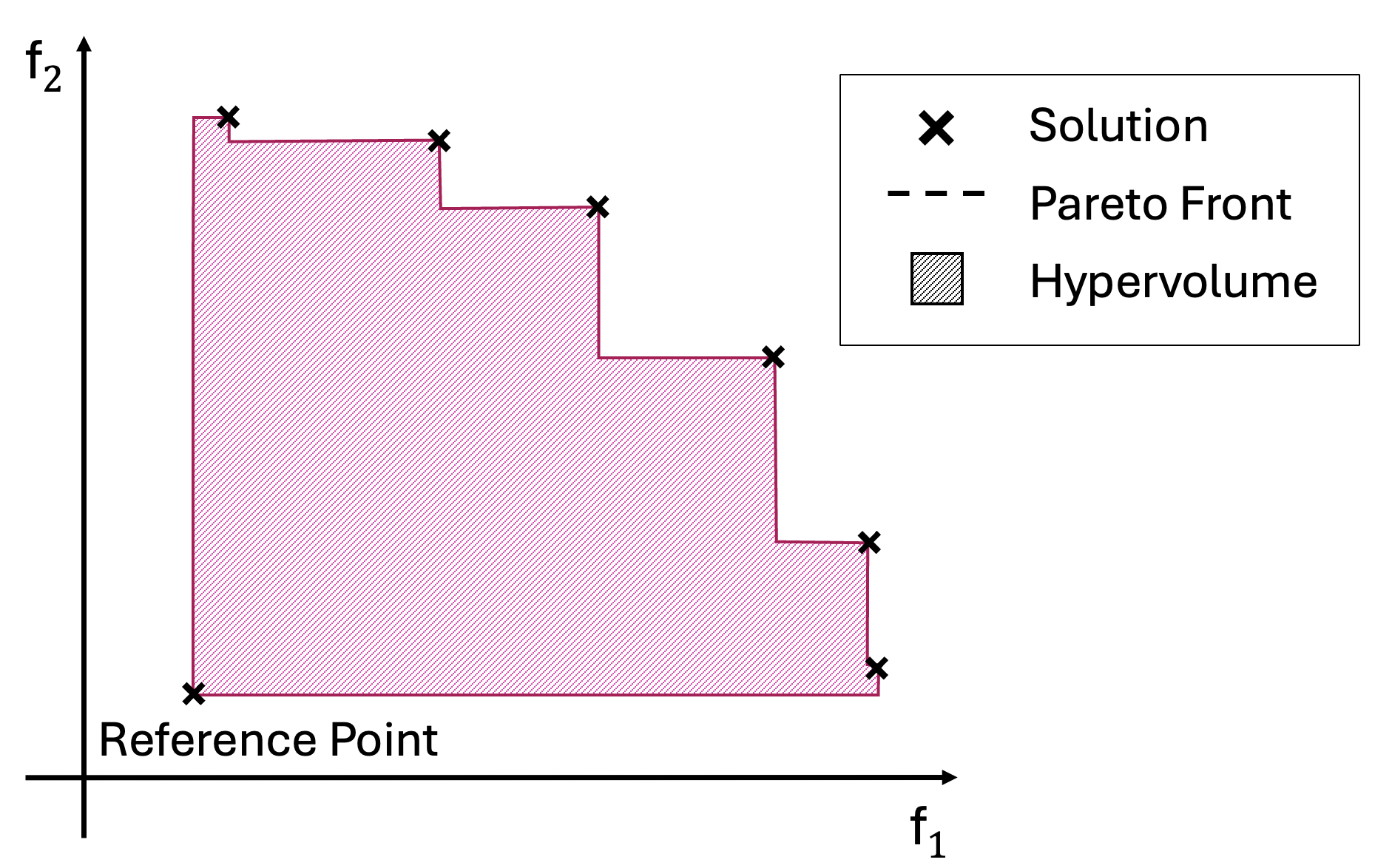}
    \caption{The Pareto Front represents the set of different trade-offs on objectives. The hypervolume reflects the area between points on the Pareto Front and a reference point.}
    \label{fig:pf}
\end{figure}

% \begin{itemize}
%     \item Multi-objective optimisation involves optimising $m$ objectives: s $\textbf{F}=f_1, f_2, ... f_m$
%     \item Definition of Pareto Dominance
%     \item Pareto front = set of possible trade-offs
%     \item Measure performance of MOO via the 
% \end{itemize}

Multi-objective optimisation considers the problem of maximising a set of different, and often conflicting, objectives at once. 
Concretely, given a solution $x \in \mathcal{X}$ and an objective $\textbf{f}: \mathcal{X} \rightarrow \mathbb{R}^k$, the problem is formalised as $\max\limits_{x \in \mathcal{X}}(f_1(x), ..., f_k(x))$.

In reality, for most problems, achieving a solution that maximises all objectives simultaneously is often unattainable. 
Consequently, the aim becomes identifying the set of solutions which achieve the best possible trade-offs on the set of objectives, known as the \textit{Pareto front} (see \Cref{fig:pf}).
%wherein each solution optimises for a different trade-off of the objectives. 
%The collection of solutions that achieve 
%Figure \ref{fig:pf} illustrates the Pareto front given the two objectives represented on each axis. 
A solution belongs to the front if it scores at least as high as the other solutions on the front on all of the objectives, and outperforms them on at least one objective \cite{guidetomorl}.
Solutions that belong to the front are called \textit{Pareto optimal}.

%A solution belongs on the front if it Pareto dominates all possible solutions for a given weighting of objectives.
%This notion of Pareto domination is used to compare solutions during the optimisation procedure.
The performance of multi-objective optimisation is often assessed via the hypervolume metric which reflects the area of space between points on the front and a fixed reference point $r$ \cite{usinghypervolumes} (see \Cref{fig:pf}). 
Pareto Fronts that contain solutions that are more evenly distributed and higher-performing on each objective will have a higher hypervolume \cite{guidetomorl}.

\subsection{Multi-Objective Quality-Diversity}\label{section:moqd}

% \begin{itemize}
%     \item MOME motivation: find diverse solutions across $m$ objectives $\textbf{F}=f_1, f_2, ... f_m$
%     \item Algorithm flow: Same as Map-Elites but store a Pareto Front in each cell
%     \item Need to state that to control storage costs and allow for parallelisation, previous works in MOQD use a fixed maximum size of Pareto Front \cite{mome, mome-pgx}.
%     \item Aim: maximise MOQD score
%     \item We note that there have been other works in \moqd that have built upon the \mome algorithm but they have been applied to policy learning in robotics environments  \cite{mome-pgx, c-mome}. To best of our knowledge, no other \moqd works have been applied to \csp. 
% \end{itemize}

Multi-Objective Quality-Diversity (\moqd) aims to generate a diverse set of high-performing solutions that maximise a combination of $k$ different objectives $\textbf{f}= [f_1, f_2, ... f_k]$.
Multi-Objective MAP-Elites (\mome) \cite{mome}, a well-established \moqd algorithm, achieves this by storing a Pareto Front of solutions in each grid cell. 
The algorithm flow closely matches that of MAP-Elites with one crucial distinction: after evaluating a new solution and identifying its corresponding cell based on its feature, the solution is placed into the cell only if it Pareto-dominates the other existing solutions in that cell.
The objective of \moqd algorithms is to find an approximation of the Pareto Front in each cell that has the best set of trade-offs and therefore maximum hypervolume \cite{mome, c-mome, mome-pgx}.
Hence, the performance of \moqd algorithms is assessed by the sum of the hypervolumes of all the Pareto Front approximations in the grid, referred to as the \moqdscore.
This objective translates to finding the best possible trade-offs across the set of objectives for each feature of interest. 

Since the Pareto Front can contain infinitely many solutions, in order to control the memory costs and allow for parallelisation, previous works in MOQD use a fixed maximum size for the Pareto Front \cite{mome, mome-pgx}.
Prior works in the field of \moqd build upon the \mome algorithm and have applied it to policy learning robotics tasks and environments. However, to the best of our knowledge, no other \moqd works have been applied to \csp. 

\section{Related Works}
\subsection{Crystal Structure Prediction}\label{section:csp}

%% CSP can be done by deep learning, but require good databases and hard to extrapolate
Within the field of Crystal Structure Prediction, recent advancements have been heavily influenced by the growing availability of crystallographic databases \cite{materialsproject}. 
For example, data-mining methods analyse these large databases with the aim to uncover insights for predicting new structures \cite{curtarolo2003predicting}.
Deep learning methods, on the other hand, harness the power of neural networks to extract features from high-dimensional crystallographic data \cite{ryan2018crystal}. 
Additionally, generative models such as auto-encoders \cite{court20203} and generative adversarial networks \cite{kim2020generative} have emerged as a creative means to produce entirely new crystal structures.
While these learning-based approaches leverage extensive data sources, they may also be biased towards these databases and therefore could be potentially limited for the discovery of entirely novel crystal structures.

%% Therefore use EAs because they require no domain knowledge
Other efforts in \csp have been driven by evolutionary algorithms (EAs) \cite{oganov2011evolutionary}.
These algorithms work by generating new structures through genetic-like operators and gradually improving the structures over multiple iterations by retaining those which are most promising.
Most of the work in applying EAs to \csp has been focused on injecting domain knowledge into the evolutionary process. 
For example, some works explore using alternative crystal representations to reduce the complexity of the search space \cite{lcoms, mends} while others introduce specialised variation operators which, for example, preserve crystal symmetries \cite{uspex, gnome, xtalopt}.

%% Some use surrogate models
A key challenge of using EAs for \csp arises from the costly nature of evaluating crystal structures. 
Therefore some methods aim to optimise a surrogate energy model rather than rely on resource-intensive DFT calculations.
While these models provide fast estimates of the energy landscape, they come with inherent reliability concerns.
However, recent works have greatly enhanced the accuracy of surrogate models by training them to be conservative in their estimates \cite{lcoms} or leveraging ensemble techniques \cite{gnome}.

\subsection{Quality-Diversity for Crystal Structure Prediction}

%% Only focus on single optimum, QD algorithms
Despite the progress made in \csp, there remains a notable gap in the CSP literature.
Most existing methods primarily focus on finding the global minimum of the energy function, largely overlooking the exploration of diverse crystal structures and their associated properties.
In a recent study by Wolinska et al. \cite{qd4csp}, the Map-Elites algorithm (see section \Cref{section:qd}) was applied to \csp, resulting in the discovery of a diverse range of crystal structures, including both known materials and promising new configurations. 
While this represents a significant advancement, this work only focused on finding stable crystal structures and disregarded other objectives for properties of the material, such as the toughness and hardness.
On the other hand, other works have used multi-objective evolutionary algorithms \cite{mends} to find such trade-offs, but do not include specific diversity-maintenance mechanisms which facilitate the discovery of diverse structures.
In contrast, our approach utilises Multi-Objective Quality-Diversity algorithms, allowing us to not only uncover diverse crystalline materials but also explore the associated trade-offs and properties within these crystal structures.

\section{Method}

\begin{figure}[h!]
    \centering
    \includegraphics[width=\linewidth]{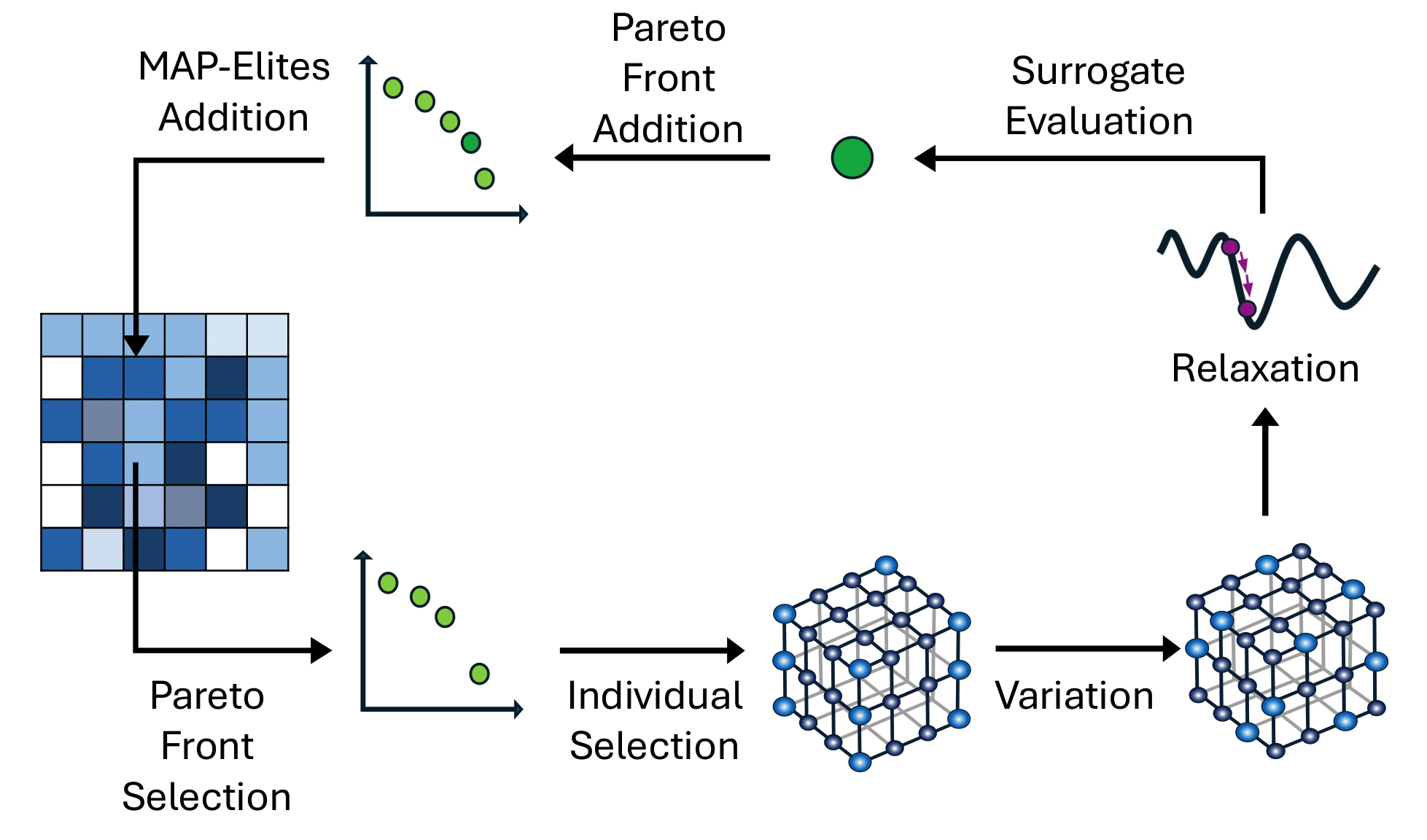}
    \caption{\momex for \csp method overview. At each iteration, solutions are selected from the grid and undergo domain-specific variation. We then perform a fixed number of relaxation steps. The new solution is evaluated via a surrogate model and added back to the archive if it belongs to the Pareto Front of the cell corresponding to its feature.}
    \label{fig:method}
    \vspace{-5mm}
\end{figure}

\subsection{Overview}

In this section we present our method, as visualised in \Cref{fig:method}).
Overall, our method adheres the general flow of the \mome algorithm which maintains a Pareto Front in each cell of a \mapelites grid: at each iteration, solutions are selected from the grid, undergo variation and are added back to the grid if they belong to the corresponding Pareto Front. 
However, we also utilise the crowding-based selection and addition exploration mechanisms from \pgx \cite{mome-pgx}, hence we term the approach \momex.
These crowding-based mechanisms are employed in order to improve exploration and ensure the maintenance of different trade-offs (see \Cref{section:crowding}).

In order to apply \momex to the \csp problem, following Wolinksa et. al \cite{qd4csp}, we inject domain knowledge into the evolutionary process.
Further details are provided in subsequent sections.

% Multi-Objective Quality-Diversity to the \csp problem (. 
% The fitness of solutions is estimated via a surrogate model (see \Cref{section:evaluation}).
% which we detail in the subsequent sections.

\subsection{Initialisation}\label{section:initialisation}
As explained in \Cref{section:csp}, crystal structures can be defined by their \textit{unit cell}, which corresponds to the smallest repeating configuration of atoms. 
Therefore, many works in \csp use the \textit{unit cell parameters} as the genotype within the evolutionary process \cite{uspex, qd4csp}.
The unit cell parameters comprise: 3 lattice parameters $a, b, c$ which describe the length of the bounding box of the unit cell, 3 parameters $\alpha, \beta, \gamma$ which describe the angles of this bounding box and  3 $\times (N - 1)$ parameters that describe the $(x, y, z)$ co-ordinates of each of the atoms in the unit cell, where $N$ is the number of atoms in the unit cell.
However, one significant challenge in \csp is the dimensionality of the search space: there are approximately $10^N$ distinct structures for a structure with  $N$ atoms.  
Of these arrangements, many candidates may not achieve the intricate symmetrical properties that crystal structures must obey for stability.
Therefore, if we were to initialise the population of structures randomly, it is highly likely that this would result in highly unstable structures which would be hard for the optimisation process to bootstrap from.
To address this, evolutionary algorithms may prefer to generate an initial population using domain-specific heuristics to ensure that the initial candidates are realistic structures \cite{qd4csp, uspex}.
We choose to use the {\fontfamily{qcr}\selectfont pyxtal} package \cite{pyxtal} to create initial candidates. 
The {\fontfamily{qcr}\selectfont pyxtal} package generates structures that conform to the permissible symmetries within a specific crystal system.

\subsection{Crowding-based Exploration}\label{section:crowding}

After generating the initial candidate structures, the algorithm enters the main algorithmic flow: at each iteration solutions are selected from the grid, mutated and evaluated for potential re-addition to the \mapelites grid.
In this work, we incorporate crowding-based selection and addition mechanisms, first introduced in \pgx \cite{mome-pgx}.
At each iteration, after selecting a Pareto Front from the grid with uniform probability, a solution from this front must be selected for variation.
The crowding-based selection operator biases the selection of solutions which lie in sparser regions of the Pareto Front which encourages exploration across all objectives.
Similarly, the crowding-based addition mechanism is used to encourage a uniform spread of solutions on the Pareto Front of each cell.
In particular, since each cell's Pareto Front has a fixed maximum size, if a new solution is added to the front of a cell which is already at maximum capacity another solution must be replaced.
In the \mome algorithm, a random solution from the front is replaced, which could result in a loss of coverage of the objective space and a decrease in \moqd score.
By contrast, the crowding-based addition operator ensures that the solutions which would provide the most even distribution across the Pareto Front are kept in each cell, ensuring that the best set of possible trade-offs are achieved. 
We refer the interested reader to the \pgx paper \cite{mome-pgx} for more details about these operators.
% To validate this design choice, we include the results of an ablation which uses uniform selection and addition mechanisms \textcolor{red}{in} \Cref{section:crowding_ablations}.
\subsection{Mutations}\label{section:mutations}
% The inherent stability of crystals relies heavily on maintaining their symmetrical properties.
% Consequently, 
The use of traditional genetic algorithm \ga variation operators poses a significant challenge in \csp since many of these operators have the potential to disrupt these symmetrical properties of crystal structures.
To address this issue, numerous works in \csp employ domain-specific mutation operators \cite{gnome, uspex, qd4csp, mends}, which are tailored to preserve these essential symmetries.
In our research, we adopt two domain specific variation operators from the Atomic Simulation Environment, {\fontfamily{qcr}\selectfont ase} package \cite{ase}: 1) \textit{strain} mutations, capable of compressing or expanding crystal structures, and 2) \textit{permutation} mutations, which facilitate the rearrangement of atoms of distinct types within the structure.
While alternative variation operators are available \cite{ase, mends, gnome, xtalopt}, an extensive study of alternative choices is beyond the scope of this work.
Moreover, previous studies have demonstrated that these selected operators are effective, especially when initiated from well-structured starting points \cite{uspex}.

\subsection{Relaxation}\label{section:relaxation}
Despite the use of domain-specific variations operators tailored to \csp, there remains the potential to inadvertently disrupt the structural crystal symmetries when applying mutations.
To mitigate this challenge, after the application of mutation operators, we apply \textit{relaxation} \cite{relaxation}.
As illustrated in \Cref{fig:csp}, relaxation is a form of local optimisation, leveraging the energy function's gradient to guide the solution towards its nearest local optima while preserving critical symmetrical properties.
The gradient of the energy function represents the force on each of the atoms and minimising the force by moving toward a local optimum leads to a more stable structure.

Relaxation is a highly effective procedure for preventing the generation of redundant crystal structures, making it a prevalent practice in \csp methodologies \cite{qd4csp, uspex, mends, gnome, oganov2011evolutionary}.
However, since relaxation is a form of local optimisation it requires performing additional energy function evaluations.
Therefore, this process comes with a computational cost and reduces the overall budget of evaluations typically available in evolutionary algorithms.
To manage the computational budget, we apply a fixed number of relaxation steps to each offspring solution.
% Similar filtering is done in other works \cite{uspex}. 
Then we use a surrogate model to estimate the force on the atoms of the offspring solution and apply a loose filtering system: if the force is still very high after these relaxation steps, it is removed from the offspring candidates.
%Then we apply a loose filtering system such that if the gradient of the energy function is still very high after these relaxation steps, it is removed from the offspring candidates.
%In other words, if the force on its atoms is high and therefore the structure is unstable, we do not consider it for addition to the archive.
% We include the results for a version of our method where we do not perform relaxation in \Cref{section:relaxation_ablations} and we demonstrate that our filtering strategy improves the discovery of real-life structures \Cref{section:thresholding_ablations}.

\subsection{Evaluating Solutions}\label{section:evaluation}
Evolutionary algorithms typically require evaluating many thousands of solutions at each iteration.
In the context of \csp, accurate energy calculations could be achieved through  DFT. However, the sheer volume of evaluations typically needed in evolutionary algorithms makes this approach prohibitively expensive.
To address this challenge, our method employs pre-trained neural networks \cite{chgnet, matgl, megnet1, megnet2}, trained on extensive materials databases as surrogate models for evaluating the fitness and features of solutions.
Although these models are less accurate compared to DFT, we contend that this limitation is manageable for several reasons.

Firstly, recent advancements in deep learning have demonstrated a trend towards improving neural network accuracy in various fields, including CSP.
Indeed, recent works which employ techniques such as ensemble training \cite{gnome} and conservative loss functions \cite{lcoms} have already facilitated enhanced surrogate model accuracy.
Thus, we anticipate that these models will continue to improve and will soon be a competitive alternative to DFT calculations.

Secondly, we hope that future work could extend our method as a tool for fine-tuning surrogate models through active-learning strategies, following a similar approach to GNoME \cite{gnome}. 
For example, we found that the surrogate model predicted some solutions to have a negative fitness, which is not feasible in real-life.
In this work, we simply filtered out these unrealistic solutions.
However, future works could re-evaluate these out-of-distribution solutions via DFT and then using them as training examples to further refine the surrogate model.
While this lies beyond the scope of our present work, it offers an intriguing avenue for future \csp research.

\section{Experimental Set Up}\label{section:experiments}
\begin{figure*}[ht]
    \centering
    \includegraphics[width=\textwidth]{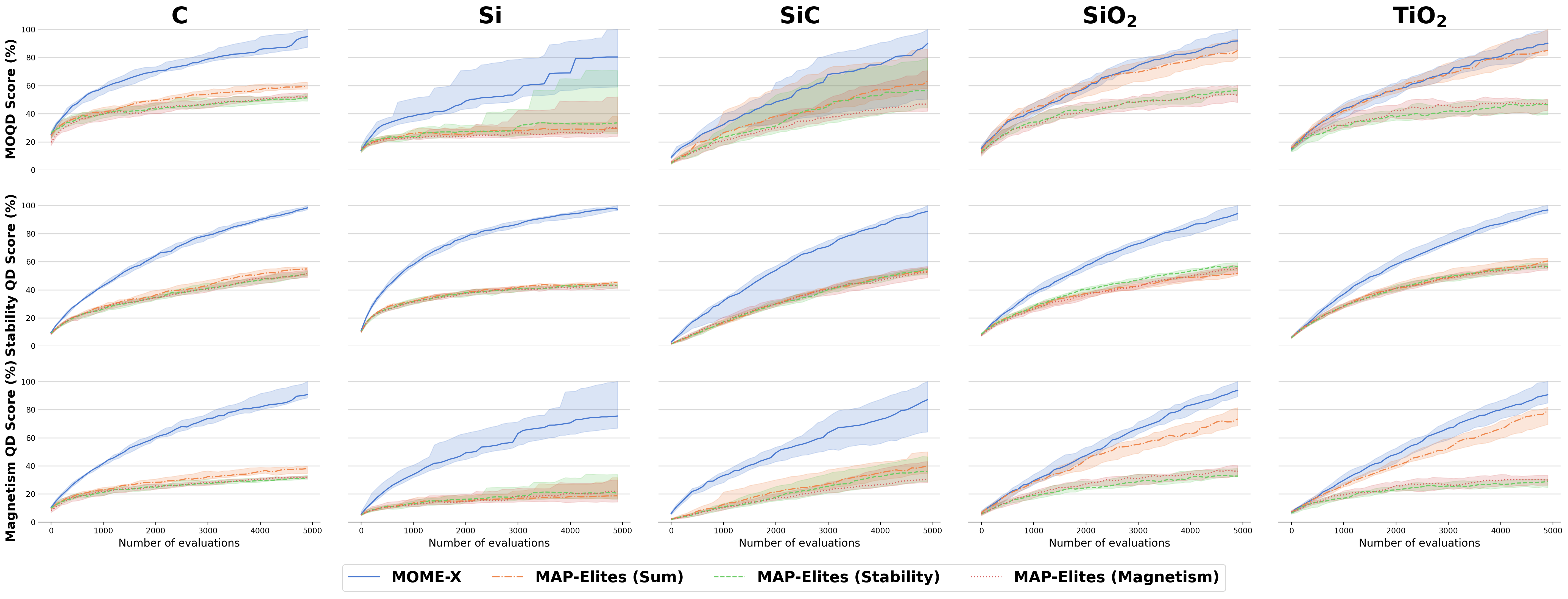}
    \caption{Median performance of \replications seeds, the shaded regions show the inter-quartile range.}
    \label{fig:moqd_results}
\end{figure*}

\subsection{Experiments}
We evaluate our method on 5 crystal systems: Carbon (C), Silicon (Si), Silicon Carbide (SiC), Silicon Dioxide (SiO$_{2}$) and Titanium Dioxide (TiO$_{2}$).
For each of the systems, we use the following features: 

\begin{enumerate}[leftmargin=*]
    \item The \textit{band gap} \cite{bandgap} reflects the energy required to move an electron from the valence band to the conduction band and therefore is an indication of the conductivity of a material. We estimate the band gap via the {\fontfamily{qcr}\selectfont "MEGNet-MP-2019.4.1-BandGap-mfi"} model from the Materials Graph Library, {\fontfamily{qcr}\selectfont matgl} \cite{matgl}.
    \item The \textit{shear modulus} \cite{shearmodulus} of a material quantifies its resistance to deformation under shear stress, providing information about its stiffness and ability to withstand forces. We estimate the shear modulus using the {\fontfamily{qcr}\selectfont "logG$\_$MP$\_$2018"} model from the Materials Graph Network Library ({\fontfamily{qcr}\selectfont MEGNet}) \cite{megnet1, megnet2}.
\end{enumerate}

The aim of the method is to find crystal structures which are diverse in their band gap and shear modulus, and are high-performing on a set of objectives. 
In this work, for all systems, we consider the following objectives \cite{qd4csp}:

\begin{enumerate}[leftmargin=*]
    \item The \textit{stability} of the crystal structure, which we quantify via the  energy function, is estimated by {\fontfamily{qcr}\selectfont CHGNet} \cite{deng2023chgnet}. The objective is to minimise the total energy of the crystal structure, reflecting the configuration that represents the most stable state for the crystal. The stability of a structure is actually assessed by its value on the energy function, with more negative values being more stable. For simplicity, we take the negative energy as fitness and seek to maximise this.
    \item We also use {\fontfamily{qcr}\selectfont CHGNet} to estimate the \textit{magnetic moment} of the structures. The magnetic moment is a vector quantity that characterises magnetic moment associated with each atom in a crystal lattice. We combine the moments for each atom into a single magnetism score, known as the total magnetic moment. For simplicity, we refer to the total magnetic moment as \textit{magnetism} henceforth.
\end{enumerate}

We note that the choice of objectives in this work serve as a starting point for applying \moqd to \csp. 
Future research avenues could explore additional material properties such as  toughness and hardness \cite{mends}.
Alternatively, we might consider using the uncertainty of an ensemble of the surrogate model predictions as a meta-objective, enabling optimisation for both stability and the confidence in that stability prediction. 
These possibilities illustrate the flexibility and potential for expanding the objectives within the \moqd framework to enhance crystal structure discovery.

\subsection{Baselines}
We evaluate our method, \momex, against three \mapelites baselines: 1) \meenergy 2) \memagnetism and 3) \mesum.
Each of these baselines corresponds to applying the \mapelites algorithm to \csp, optimising for stability, magnetism and an additive sum of the two respectively. 
We use the same mutation operators, initialisation and relaxation procedures for all of the baselines and \momex.
However, it should be noted that a straightforward comparison between \mome algorithms and  \mapelites algorithms is non-trivial, since \mapelites algorithms store at most one solution per cell whereas \mome algorithms can store several.
In particular, if we use an archive tessellated into $c$ cells for \momex, each with a maximum Pareto Front length $p$, \momex could have a maximum population size of $c \times p$.
Therefore, for fairness, in all \mapelites baselines we tessellate the archive into  $c \times p$ cells.

On the other hand, calculating \moqd metrics on grids with different tessellation sizes would also not be a fair comparison. 
Therefore, to report metrics for the baselines,  we also maintain passive \mome archives alongside the main \mapelites archives \cite{mome, mome-pgx}. 
At each iteration, all of the solutions from the \mapelites archives are added to these passive \mome archives using the normal Pareto Front addition rules. 
The passive archives are then used to calculate and report metrics, but do not interact with the main algorithm flow.

\subsection{Hyperparameters}
For \momex, we use an archive with a CVT tessellation \cite{cvt} of 200 cells, each with a maximum Pareto Front length of $10$.
Therefore, for \mapelites baselines we use a CVT tessellation of 2000 cells.
We run all experiments for a total of $5000$ evaluations, selecting $100$ solutions from the grid at each iteration.
For all systems, we consider structures with 24 atoms, with an initial unit cell volume of 450 A$^3$ and ratio of covalent radii of 0.4.
For multi-atom systems (SiC, SiO$_{2}$, TiO$_{2}$) we apply strain and permutation mutation operators with equal probability and for single-atom systems (Si, C) we only apply strain mutation operators.
After applying the variation operators we use 100 relaxation steps and filter out solutions with a force greater than $1.0$ eV/A.
We use a reference point of $[0, 0]$ to calculate the hypervolume.
Each experiment was repeated for \replications replications.

\section{Results}
\begin{figure*}[h!]
    \centering
    \includegraphics[width=0.9\textwidth]{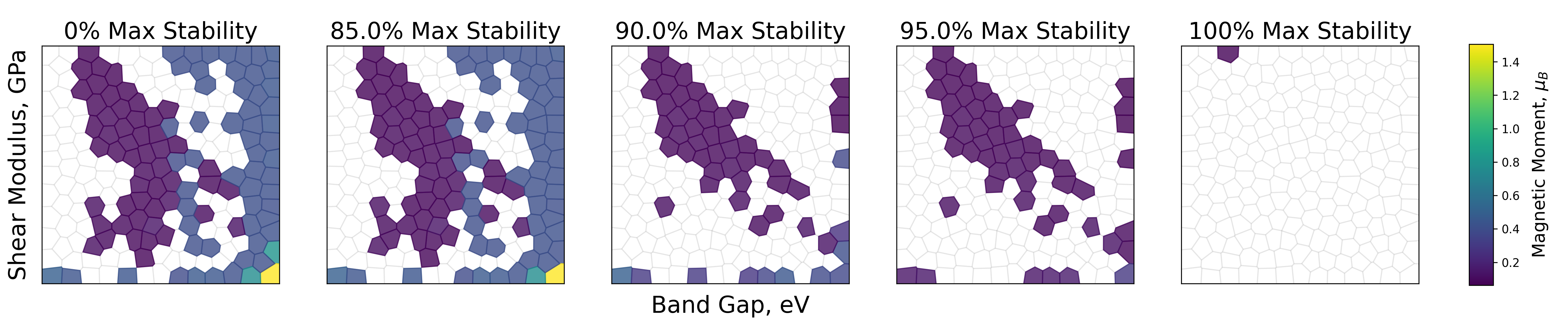}
    \caption{Archive plot for Silicon Carbide, colour coded to show the maximum magnetism fitness in each cell. Different plots show different threshold levels for the minimum stability of solutions.}
    \label{fig:illumination}
\end{figure*}

\subsection{Multi-Objective Quality-Diversity Results}\label{section:moqd_results}

We first evaluate \momex against the \mapelites baselines using the following \moqd metrics:

\begin{enumerate}[leftmargin=*]
    \item \moqdscore: the sum of the hypervolumes of each Pareto Front of the \mome archive. This metric aims to reflect whether our approach is able to find solutions which are diverse in the feature space, that achieve different trade-offs on each of the objectives (see \Cref{section:moqd}).
    \item \energyqdscore: the sum of the energy scores of all solutions in the archive. This is what the \meenergy baseline is explicitly trying to maximise. 
    \item \magmomqdscore: the sum of the magnetism scores of all solutions in the archive. This is what the \memagnetism baseline is explicitly trying to maximise. 
\end{enumerate}

We verify the statistical significance of our results by reporting the $p$-value from a Wilcoxon signed-rank test \cite{wilcoxon1992individual} using a Holm-Bonferroni correction \cite{holm_bonf}.

Firstly, \Cref{fig:moqd_results} shows that \momex outperforms or matches the performance of all of the baselines across all of the metrics in all of the systems.
Crucially, this shows that \momex is able to find the largest selection of possible trade-offs of stability and magnetism for varying band-gaps and shear-moduli.
Interestingly, \momex outperforms ($p < 0.03$) all baselines on the \moqdscore for single-atom systems (C, Si) but matches the performance of \mesum on multi-atom systems (SiC, TiO$_{2}$, SiO$_{2}$). 
This suggests that scalarisation of the objectives may not always achieve a wide variety of possible trade-offs.

\Cref{fig:moqd_results} shows also that \momex also outperforms ($p < 0.03$) all other baselines on the \energyqdscore and \magmomqdscore.
This is a particularly intriguing result, given that these metrics are what \meenergy and \memagnetism are respectively explicitly aiming to maximise.
This corroborates previous observations that simultaneously optimising over several objectives may provide helpful stepping stone solutions that aid exploration and, in turn, can help to maximise each objective separately \cite{mome-pgx}.
Additionally, we observe that \momex also outperforms \mesum on these metrics.
This could suggest that \mesum succeeds in finding solutions that are balanced across the two objectives, but is less able to find solutions that are high-performing on each objective separately and therefore provides fewer possible trade-offs for each cell.

Finally, we note two further metrics that can be used to asses \moqd performance: the \coverage and the \globalhypscore \cite{mome, mome-pgx}.
The \coverage is the proportion of cells in the archive that have at least one solution and aims to assess whether the algorithm can find solutions that are diverse in the feature space.
The \globalhypscore is the hypervolume of the Pareto Front of all the solutions from the entire archive and aims to assess the multi-objective performance of the algorithm, when disregarding the solutions' features.
We find that these metrics are similar for most algorithms so report the results in \Cref{section:extra_appendix}.
We note that this result is expected since all baselines are designed to find solutions cover the feature space and the global Pareto Front is attained by most methods.
However, we emphasise that only \momex was able to solve the combinatorial problem of finding the best Pareto Front for each feature, as reflected in its high \moqdscore.

\begin{figure*}[h!]
    \centering
    \includegraphics[width=\textwidth]{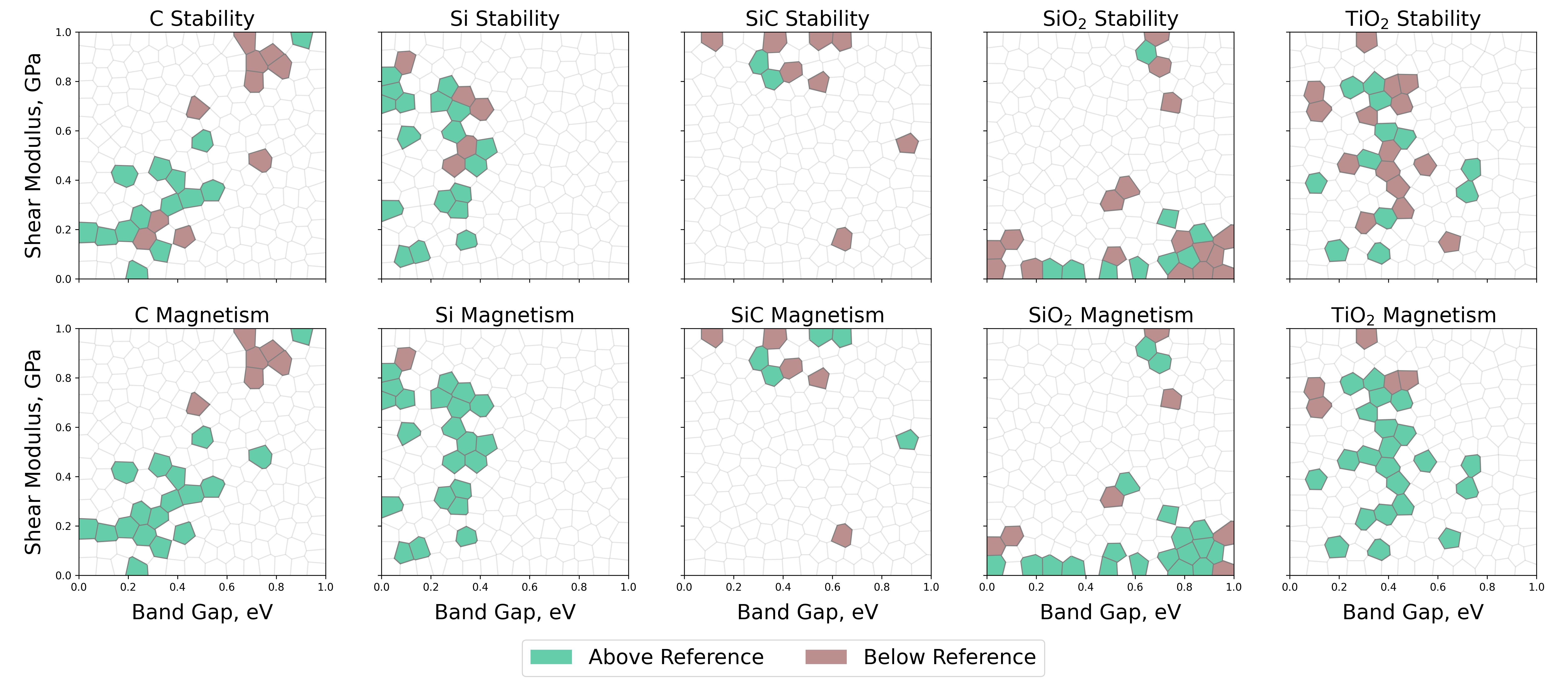}
    \caption{Archive visualisation with cells colour-coded to indicate whether \momex found a solution in the same cell as the reference solution that was higher-performing. Top row: stability objective. Bottom row: magnetism objective.}
    \label{fig:reference_illumination}
\end{figure*}
\subsection{Illumination}

In \Cref{section:moqd_results} we demonstrated that \momex surpasses all \mapelites baselines across various \moqd metrics, highlighting its ability to discover an extensive collection of diverse crystal structures which achieve distinct trade-offs of stability and magnetism.
However, we recognise that in many \csp applications, the importance of these objectives may not be equal.
Therefore, in order to further understanding what trade-offs are possible, in this section we introduce a visualisation method that portrays the levels of magnetism that are attainable for varying degrees of stability.

To construct this representation, we first computed the minimum and maximum stability scores within the entire archive of one replication. 
Then, we visualise the archive, colour-coded according to the highest magnetism score that is achieved by solutions for varying degrees of interpolation between this minimum and maximum stability. 
For example, a $90\%$ interpolation shows the best possible magnetism scores in each cell for solutions which achieve at least $90\%$ of the maximum stability of the entire archive.

\Cref{fig:illumination} shows an example of this plot for the \momex run that achieved the median \moqd score for Silicon Carbide (visualisations for other crystal systems are provided in \Cref{section:illumination_appendix}).
This visualisation reveals when the minimum threshold for stability is lower, the magnetism of solutions is higher.
This highlights that there is a genuine trade-off between the stability and magnetism of solutions.
However, \Cref{fig:illumination} also shows there are many solutions that perform highly on both objectives, given that there are still many solutions with high magnetism scores, even at high interpolation levels of $85\%$, $90\%$, and $95\%$.
%For example, we can see that many of the solutions with the highest magnetism are still above the $85\%$ threshold of the maximum stability of the entire archive.
Therefore, if the end-user is willing to sacrifice some level of stability for a highly magnetic material, \Cref{fig:illumination} suggests that there are crystal structures that would satisfy this.
We emphasise that this visualisation method, while used to illustrate trade-offs between stability and magnetism, could be adapted for exploring trade-offs with other objectives (as detailed in \Cref{section:experiments}) and finding solutions that achieve targeted properties.
\vspace{-10pt}
\subsection{Evaluation Against Reference Structures}

While utilising \moqd metrics offers valuable insights into the performance of \momex, we also wish to validate that our approach is useful for the discovery of real-life crystal structures that could have practical use in materials science applications.
This approach is pivotal in confirming that our method is not just yielding solutions that deceive the neural-network surrogate models but rather uncovering genuinely realistic configurations.
To accomplish this, we cross-reference the structures found by each of the baselines against those contained in the \textit{Materials Project Database} \cite{materialsproject}, which is an extensive database of approximately 150,000 materials.
To verify whether a solution found by an algorithm matches a real-life reference structure, we use the  {\fontfamily{qcr}\selectfont StructureMatcher} method from the {\fontfamily{qcr}\selectfont pymatgen} \cite{pymatgen}.
This method provides a series of checks on the unit-cell parameters to determine if there is a match.
We consider a solution to match the reference structure if it 1) is a match according to the {\fontfamily{qcr}\selectfont StructureMatcher} and 2) lies in the same centroid as the reference structure.

\begin{table}[ht]
\centering
  \caption{Median number of reference structure matches}
  \scalebox{0.9}{\begin{tabular}{c|c|c|c|c}
    \toprule
    & \makecell{\momex}
    & \makecell{\meenergy}
    & \makecell{\memagnetism}
    & \makecell{\mesum} \\

    \midrule

    C
    & \textbf{3.0}
    & 2.0
    & 2.0
    & 2.0 \\
    
    \midrule

    Si
    & \textbf{6.0}
    & 2.0
    & 3.0 
    & 3.0 \\

    \midrule
    
    SiC
    & 0.0
    & 0.0 
    & 0.0
    & 0.0 \\
    
    \midrule
    
    SiO$_{2}$
    & \textbf{7.0}
    & 1.0 
    & 3.0
    & 3.0 \\

    \midrule
    TiO$_{2}$
    & \textbf{4.0}
    & 2.0
    & 2.0
    & 1.0 \\    
    
    \bottomrule
    
\end{tabular}\label{tab:structure_matcher_results}}

\end{table}

\Cref{tab:structure_matcher_results} shows the median number of reference structure matches found by each of the algorithms.
Interestingly, we find that \momex is able to find more real-life structures than other baselines. 
We hypothesise that this could be because optimising over several objectives provides diverse stepping stone solutions that can aid exploration and thus improve the discovery of realistic structures.

While it is promising to see that \momex is able to re-discover real-life structures, we also wanted to verify whether any of the other solutions found by the algorithm outperform these reference structures, on either of the objectives.
\Cref{fig:reference_illumination} shows the archives of the runs that score the median \moqdscore in each of the systems.
The cells of the reference structures are colour coded according to whether, for each structure, \momex finds a solution in the same cell that outperforms the reference structure on either of the objectives. 
In other words, \Cref{fig:reference_illumination} visualises whether \momex is able to find solutions that have similar features to the reference solutions, but are either more stable or have higher magnetism.
Excitingly, these results show that \momex finds many promising structures that could be better than the known reference materials. 

While we note that it is likely that some of these structures could score highly due to inaccuracies of the surrogate model, we argue that this presents an opportunity for future research, rather than a limitation of our method. 
Indeed, as explained in \cref{section:evaluation}, we believe that finding solutions which are able to fool the surrogate model lays the foundation for a very interesting line of future research which use these types of solutions as training examples to improve surrogate models in \csp via active learning. 
This process could be employed in an adversarial loop:  surrogate models could aid the discovery of novel crystal structures and then these structures could be re-assessed via DFT calculations and used to create even better surrogate models. 
In summary, \Cref{fig:reference_illumination} shows numerous solutions that are either a) higher-performing than known materials or b) could be used to improve surrogate model accuracy, both of which are advantageous outcomes.

\section{Conclusion}
In this work, we applied Multi-Objective Quality-Diversity to Crystal Structure Prediction.
Using this approach, we were able to find a range of crystal structures that achieved different trade-offs of magnetism and stability, and we proposed a method for visualising these trade-offs.
We found that \momex discovered more real-life structures than \mapelites baselines and also had a higher performance on the \energyqdscore and \magmomqdscore, suggesting that optimising over many objectives may improve exploration.
Finally, we explained how our method could fit into a larger framework for active learning, and could be used to optimise for other objectives such as toughness \cite{uspex} or meta-objectives such as surrogate model uncertainty.
Future work could consider improving the efficiency of our method by using gradient-based mutation operators \cite{mome-pgx, dqd}.
Alternatively, we could apply our approach to similar problems in other fields, such as the discovery of protein folds \cite{alphazero_db}.

\section*{Acknowledgements}
This work was supported by PhD scholarship funding for Hannah from InstaDeep.

\clearpage

\bibliographystyle{ACM-Reference-Format}
\bibliography{main}

\clearpage
\onecolumn
\appendix
\section{Supplementary MOQD Results}\label{section:extra_appendix}

\begin{figure*}[h!]
    \centering
    \includegraphics[width=0.9\textwidth]{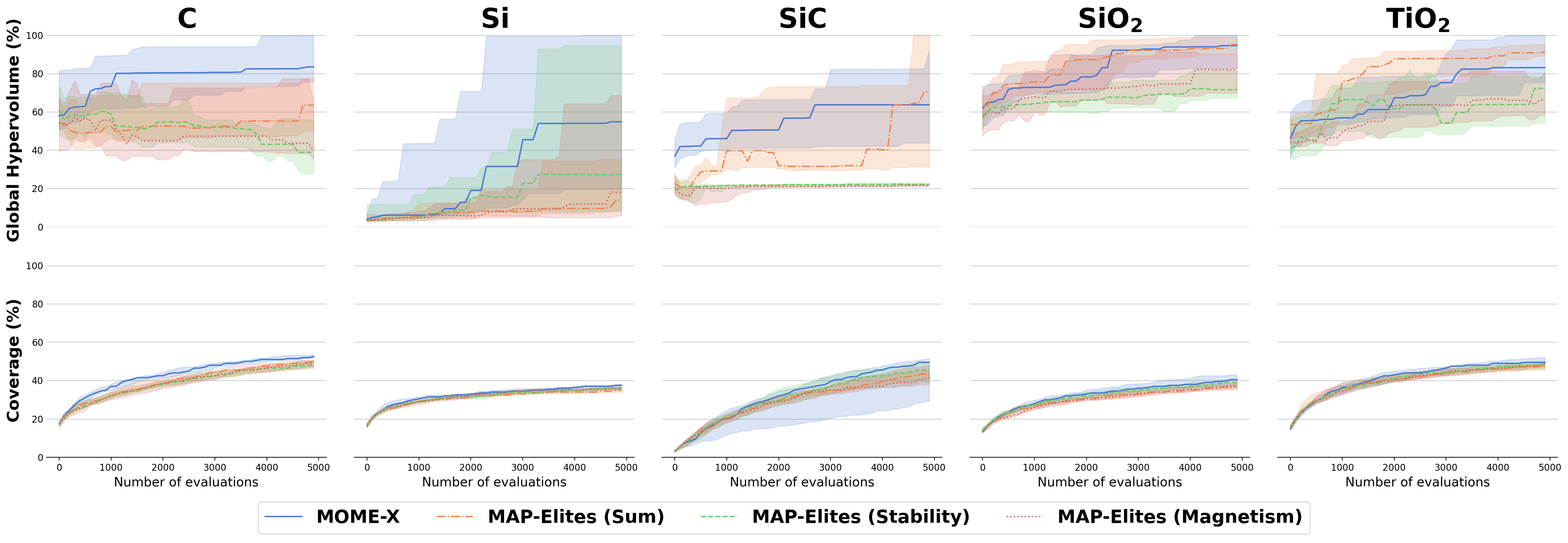}
    \caption{Median performance of \replications seeds, the shaded regions show the inter-quartile range.}
    \label{fig:extra_metics}
\end{figure*}

\section{Multi-Objective Illumination of Crystal Systems}\label{section:illumination_appendix}

\begin{figure*}[h!]
    \centering
    \includegraphics[width=0.9\textwidth]{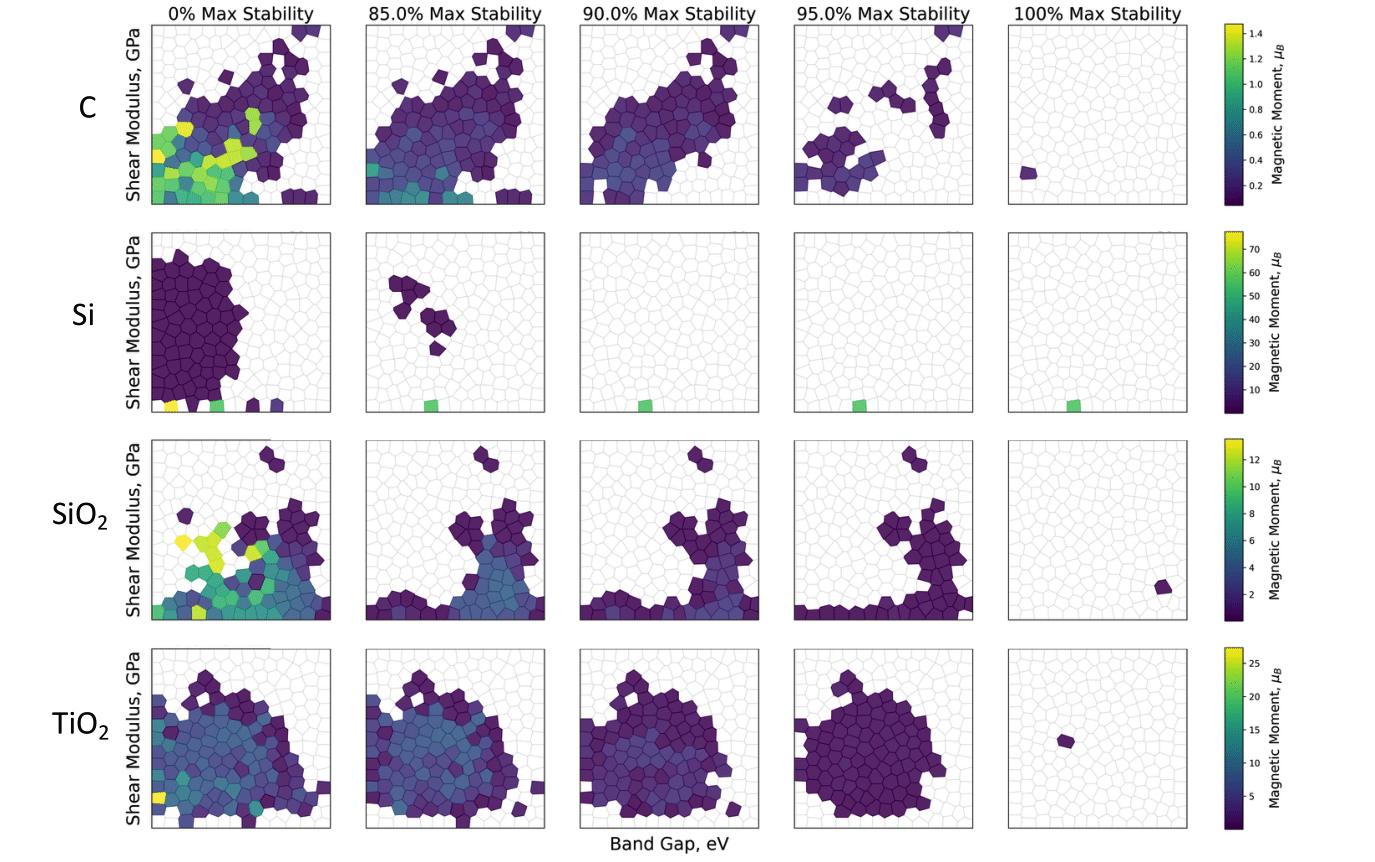}
    \caption{Archive plots, colour coded to show the maximum magnetism fitness in each cell. Different plots show different threshold levels for the minimum stability of solutions, based on the the minimum and maximum stability of the whole archive.}
    \label{fig:extra_illumination}
\end{figure*}

\end{document}